\documentclass[runningheads, a4paper]{llncs}

\usepackage{amssymb}
\usepackage{graphicx}

\usepackage{url}
\newcommand{\keywords}[1]{\par\addvspace\baselineskip
\noindent\keywordname\enspace\ignorespaces#1}

\usepackage[utf8]{inputenc}

\usepackage{hyperref}
\usepackage{cite}

\usepackage{float}

\usepackage{caption}
\usepackage{subcaption}

\usepackage{pdflscape}

\usepackage{color, soul}

\usepackage{array}
\newcolumntype{x}[1]{>{\centering\let\newline\\\arraybackslash\hspace{0pt}}p{#1}}

\usepackage{wrapfig}

\usepackage[vlined, linesnumbered, ruled]{algorithm2e}
\makeatletter
\renewcommand{\@algocf@capt@plain}{above}
\makeatother

\begin{document}

\mainmatter 

\title{Towards S-NAMO: Socially-aware Navigation Among Movable Obstacles}

\titlerunning{Towards Social NAMO}

\author{Benoit Renault$^{a,b}$ \and Jacques Saraydaryan$^{a,c}$ \and Olivier Simonin$^{a,b}$}

\institute{
$^a$ CITI Lab., INRIA Chroma,
$^b$ INSA Lyon,
$^c$ CPE Lyon,\\
Université de Lyon, Villeurbanne, France \\
}

\authorrunning{Renault \and Saraydaryan \and Simonin}

\toctitle{Towards Social NAMO}
\tocauthor{Renault, Saraydaryan, Simonin}
\maketitle

\nocite{
wilfong_motion_1991,
chen_practical_1991,
demaine_pushpush_2000,
ota_rearrangement_2004,
okada_environment_2004,
stilman_navigation_2005,
kim_traversability_2006,
stilman_planning_2007,
stilman_navigation_2007,
nieuwenhuisen_effective_2008,
stilman_planning_2008,
van_den_berg_path_2009,
kakiuchi_working_2010,
wu_navigation_2010,
levihn_efficient_2011,
levihn_navigation_2011,
levihn_hierarchical_2013,
levihn_planning_2013,
levihn_foresight_2013,
kruse_human-aware_2013,
clingerman_estimating_2014,
mueggler_aerial-guided_2014,
levihn_locally_2014,
rios-martinez_proxemics_2015,
clingerman_dynamic_2015,
castaman_sampling-based_2016-1,
castaman_sampling-based_2016,
moghaddam_planning_2016,
scholz_navigation_2016,
charalampous_recent_2017,
sun_semantic_2017,
jumel_context_2018,
meng_active_2018
}

\begin{abstract} 

In this paper, we present an in-depth analysis of Navigation Among Movable Obstacles (NAMO) literature, notably highlighting that social acceptability remains an unadressed problem in this robotics navigation domain. The objectives of a Socially-Aware NAMO are defined and a first set of algorithmic propositions is built upon existing work. We developed a simulator allowing to test our propositions of social movability evaluation for obstacle selection, and social placement of objects with a semantic map layer. Preliminary pushing tests are done with a Pepper robot, the standard platform for the \href{http://www.robocupathome.org/}{Robocup@home} SSPL\footnote{SSPL: Social Standard Platform League}, in the context of our participation (\href{https://robocup-lyontech.github.io/}{LyonTech Team}).

\end{abstract}

\section{Introduction} 

In 2005, Stilman et al. \cite{stilman_navigation_2005} formulated the field of Navigation Among Movable Obstacles (NAMO). The NAMO problem consists in planning a path from a start to a goal position, while moving obstacles if necessary. It extends the well known Piano Mover’s Problem by differentiating static and movable obstacles, and allowing the manipulation of the later if it minimizes the chosen cost function (eg. travel distance, time, energy). Contexts like service robotics or search and rescue, in particular, would definitely benefit from algorithms capable of dealing with manipulable clutter, doors or objects.

In the last two decades, the growing interest in service robotics, implying robot navigation in human-populated environments, has sparked interest in Social Robotics, and more specifically Socially-Aware Navigation (SAN) \cite{kruse_human-aware_2013, rios-martinez_proxemics_2015, charalampous_recent_2017}. Basically, it also extended the basic navigation problem: now not only must the robot find a plan that ensures physical safety (no collisions), minimizes the travel distance, time or energy, but also the disturbance to humans\footnote{In Socially-Aware Navigation, disturbance is used as synonym for 'discomfort', the feeling of being unsafe \cite{rios-martinez_proxemics_2015}.}.

Until now, to the best of our knowledge, has never been considered in NAMO problems the necessity of minimizing disturbance to humans (or any other type of autonomous agents). Thus, we want to create Social NAMO algorithms: ones that allow an autonomous agent to go from an initial pose to a goal pose, forbidding collision with obstacles but not their displacement, minimizing both robot's displacement cost (distance, time or energy) and disturbance to humans.

To achieve this, we make the following contributions: in Section \ref{analysis_section}, we provide an analysis 
of existing NAMO-related works. Then, in Section \ref{extension_section}, we define the expectations for Social NAMO and propose two extensions applied to Wu \& Levihn's approach
\cite{wu_navigation_2010, levihn_navigation_2011, levihn_locally_2014}. We introduce social movability evaluation in obstacle selection, and a semantic map layer to deal with social placement of objects.
Finally, in Section \ref{experimentations_section}, we propose
experiments based on our open simulator and the Pepper robot, in the view of our \href{http://www.robocupathome.org/}{RoboCup@Home} participation (\href{https://robocup-lyontech.github.io/}{LyonTech Team}).
We provide closing remarks and discuss future work in Section \ref{conclusion_section}.

\section{NAMO: Analysis of Existing Works}\label{analysis_section} 

The following paragraphs give an overview of NAMO through the discussion of the used world representations, notion of cost \& optimality, manipulation characteristics and finally, the actual planning algorithms that rely on them. Also, we will point out how they relate to socially-aware navigation and its constraints. A synthesis of the main comparison criteria is given in table \ref{synthesis_table}.

\paragraph{\textbf{World representation}} NAMO relies on an object-based representation of the world \cite{chen_practical_1991, okada_environment_2004, stilman_navigation_2005, stilman_planning_2007, nieuwenhuisen_effective_2008, stilman_planning_2008, van_den_berg_path_2009, kakiuchi_working_2010, wu_navigation_2010, levihn_navigation_2011, levihn_hierarchical_2013, levihn_planning_2013, levihn_foresight_2013, mueggler_aerial-guided_2014, levihn_locally_2014, castaman_sampling-based_2016, castaman_sampling-based_2016-1, moghaddam_planning_2016, scholz_navigation_2016, sun_semantic_2017, meng_active_2018} (in opposition to an occupation-space-based one): in order to chose the best obstacle placement, it is necessary to reason about them as separate entities. Final placement selection is what actually tells NAMO apart from the well-known field of Rearrangement Planning \cite{ota_rearrangement_2004}. Inspired by the works of Kim et al. on traversability affordance \cite{kim_traversability_2006}, Clingerman et al. \cite{clingerman_estimating_2014, clingerman_dynamic_2015} represent movable obstacles as high values in a costmap, but they recon that it can't be called a NAMO algorithm, since it does not allow to control obstacle placement (the robot simply tries to "go through the obstacle").

Semantic information about Movable Obstacles is key to these algorithms. The most basic need is the 'movability' attribute, in addition to the obstacles position and shape. In the literature, individual obstacles are simply assumed to have a boolean attribute of being movable or not. This attribute has until now been given as input \cite{chen_practical_1991, okada_environment_2004, stilman_navigation_2005, stilman_planning_2007, nieuwenhuisen_effective_2008, stilman_planning_2008, van_den_berg_path_2009, levihn_hierarchical_2013, levihn_planning_2013, castaman_sampling-based_2016, castaman_sampling-based_2016-1, moghaddam_planning_2016} (mainly for simulation-only algorithms), determined on-line from obstacle visual recognition results \cite{levihn_foresight_2013, mueggler_aerial-guided_2014, scholz_navigation_2016, sun_semantic_2017, meng_active_2018} or by manipulation tentative \cite{kakiuchi_working_2010, wu_navigation_2010, levihn_navigation_2011, levihn_locally_2014} (for real-world experiments). In order to be more realistic, other semantic information is used in more advanced approaches, like object kinematics and physics (mass, center of inertia, \dots), but successfully used only in simulated propositions \cite{chen_practical_1991, stilman_navigation_2005, stilman_planning_2007, castaman_sampling-based_2016, castaman_sampling-based_2016-1}, with mixed results in actual real-world implementations \cite{scholz_navigation_2016} (these characteristics are hard to determine with current robot sensing capabilities). Other types of obstacles than movable or unmovable, like humans or autonomously moving objects have never been considered in the NAMO literature: a standard hypothesis is that the robot is the only autonomous agent in the environment. We recapitulate this in the 'Movability' column of Table \ref{synthesis_table}.

We must also say that rather few NAMO propositions have been applied in a real-world setting \cite{stilman_planning_2007, kakiuchi_working_2010, levihn_foresight_2013, mueggler_aerial-guided_2014, scholz_navigation_2016, sun_semantic_2017, meng_active_2018}, and when they are, they always maintain a 3D representation of the world, though all NAMO algorithms execute their path finding subroutines in a 2D plane. 3D data is mainly used to allow for proper grasping of obstacles, but also for cross-plane rearrangement planning \cite{meng_active_2018}(e.g. pick\&place an object from ground to tabletop). Data is either acquired through external cameras and markers \cite{stilman_planning_2007, mueggler_aerial-guided_2014, scholz_navigation_2016} to position priorly known polyhedral models of movable obstacles (eg. chairs, tables, \dots), guaranteeing negligible uncertainty as to the environment's state, or by on-board sensors only \cite{kakiuchi_working_2010, levihn_foresight_2013,sun_semantic_2017, meng_active_2018}. A limited number of propositions actually are able to deal with no prior or partial geometric knowledge \cite{levihn_foresight_2013, sun_semantic_2017, meng_active_2018, kakiuchi_working_2010, wu_navigation_2010,  levihn_locally_2014}, uncertainty as to object positioning \cite{stilman_planning_2007, kakiuchi_working_2010, levihn_hierarchical_2013, levihn_planning_2013, levihn_foresight_2013, castaman_sampling-based_2016, castaman_sampling-based_2016-1, scholz_navigation_2016, sun_semantic_2017, meng_active_2018}, object movability \cite{kakiuchi_working_2010, levihn_hierarchical_2013, levihn_planning_2013, levihn_foresight_2013, scholz_navigation_2016} or object kinematics/physics \cite{levihn_planning_2013, scholz_navigation_2016} (Recapitulated in Table \ref{synthesis_table}, columns 'Prior', 'Uncertainty' and 'Real-World').

In the end, SAN and NAMO both depend on semantic knowledge in addition to spatial knowledge: the robot needs to differentiate objects, associate proper attributes with them, but also understand their relations to the whole environment. Systematic segmentation and identification of as many obstacles as possible thus appears to be a basic requirement for a Social NAMO.

\paragraph{\textbf{Cost \& Optimality}} There is a wide variety of cost functions used in NAMO: distance, time, energy, number of moved obstacles, probability of success, that are sometimes combined or used alternatively: these are synthesized in Table \ref{synthesis_table}, column 'Cost'. The choice of a cost that only takes displacement distance into account can be motivated by the hypothesis that the weight of the movable obstacles is negligible in regard to the physical capabilities of the robot. It is however evident that if manipulating an obstacle results in a significant change of speed and energy requirements compared to a sole navigation task, time and energy become way more appropriate choices.

NAMO Algorithms rarely seek completeness like \cite{stilman_navigation_2005, stilman_planning_2007, moghaddam_planning_2016, nieuwenhuisen_effective_2008, van_den_berg_path_2009}. None have achieved global optimality, and only Levihn \cite{levihn_navigation_2011, levihn_locally_2014} can claim a local optimality for a very simplified variant of the problem where a plan can only contain one movable obstacle (see Table \ref{synthesis_table}, 'Comp.' and 'Opt.' columns). This situation actually makes sense, when one knows that a simplified variation of the NAMO problem, where the robot is considered as a square, all planar obstacles as rectangles of four sizes or “L-Shaped”, parallel to the x- or y-axis, has been proved to be NP-Hard \cite{wilfong_motion_1991}, and even PSPACE-hard if the final positions are predetermined. When obstacles are further reduced to square blocks limited to translations on a planar grid, the problem still remains NP-Complete \cite{demaine_pushpush_2000}.

In SAN, the presence of humans brings strong uncertainty that prevents proving global optimality and completeness of navigation strategies. It also results in the need to take social costs into account during navigation \cite{kruse_human-aware_2013, rios-martinez_proxemics_2015, charalampous_recent_2017} to represent risk of disturbance to humans. In a Social NAMO, we must thus also take social costs into account, and extend them from the robot to the moved obstacles: other entities can now suffer the consequences and risks of a carelessly moved obstacle.

\paragraph{\textbf{Manipulation}} In \cite{stilman_navigation_2007}, Stilman formalized three main classes of obstacle manipulation procedures: Grasping (constrained contact), Pushing (constrained motion), and Manipulation Primitives (relies on forward simulation of object dynamics, translational or rotational slip may occur). According to the results exposed in Table \ref{synthesis_table}, Column 'Manipulations', grasping is the most popular class, likely because it is the most reliable. Pushing has also been considered because large objects cannot necessarily be grasped. Manipulation Primitives have also been experimented with, but real-world implementations require external cameras to work \cite{stilman_planning_2007, scholz_navigation_2016}.

In order to reduce the manipulation search space, there are 3 common strategies. The first, applied by all but \cite{chen_practical_1991}, is to consider that only one obstacle may be manipulated at once (no cascade effect on nearby movables). The second, also commonly used by all but \cite{chen_practical_1991, van_den_berg_path_2009, moghaddam_planning_2016} (which have never been applied in a real-world situation), is to consider a limited set of contact/grasping points, facilitating backward search for robot pose for manipulation. This semantically makes sense, in particular since some obstacles have specific contact points (eg. top of chair, regularly spaced points on table side, \dots) \cite{okada_environment_2004, stilman_planning_2007, levihn_foresight_2013, mueggler_aerial-guided_2014, castaman_sampling-based_2016, castaman_sampling-based_2016-1, sun_semantic_2017, meng_active_2018, stilman_navigation_2005, stilman_planning_2008, levihn_planning_2013, scholz_navigation_2016, kakiuchi_working_2010, wu_navigation_2010, levihn_navigation_2011, levihn_hierarchical_2013, levihn_locally_2014}. Finally, the third strategy is to limit manipulation to translations in specific directions \cite{stilman_planning_2007, van_den_berg_path_2009, kakiuchi_working_2010, wu_navigation_2010, levihn_navigation_2011, levihn_hierarchical_2013, levihn_locally_2014}.

In a Social NAMO, the robot should bring a particular attention to human safety and comfort. Favoring the most reliable manipulation classes when possible, and reducing the complexity of the manipulation (thus, its chance to fail in a way that may put humans or their belongings at risk) would be of circumstance.

\paragraph{\textbf{Planning algorithms}} While some solutions \cite{chen_practical_1991, castaman_sampling-based_2016, castaman_sampling-based_2016-1} propose tightly woven algorithms that do not clearly distinguish the different aspects of NAMO (iteration over movable obstacles, possible actions and path computations), we can usually tell apart a high-level decision planner and two path planning subroutines. These subroutines can loosely be identified as transit (robot only) and transfer (+obstacle) path planners.

The most proposition-specific planner is generally the high-level task planner. While some propositions are explicitly based on existing algorithms, like Dijkstra \cite{chen_practical_1991}, DFS \cite{stilman_navigation_2005, stilman_planning_2007, stilman_planning_2008, moghaddam_planning_2016}, BHPN \cite{levihn_foresight_2013}, Markov Decision Processes + Monte-Carlo Tree Search \cite{levihn_hierarchical_2013, levihn_planning_2013, scholz_navigation_2016}, KPIECE+A* \cite{castaman_sampling-based_2016, castaman_sampling-based_2016-1}, others appear to have developed their approach from scratch \cite{okada_environment_2004, nieuwenhuisen_effective_2008, van_den_berg_path_2009, kakiuchi_working_2010, wu_navigation_2010, levihn_navigation_2011, mueggler_aerial-guided_2014, levihn_locally_2014, sun_semantic_2017, meng_active_2018}, though \cite{levihn_navigation_2011, levihn_locally_2014} is based off \cite{wu_navigation_2010}, and \cite{meng_active_2018} has been inspired by \cite{wu_navigation_2010, levihn_locally_2014}. In order to reduce computation time, most high-level planners resort to ways of prioritizing the most promising obstacles but \cite{chen_practical_1991, nieuwenhuisen_effective_2008, van_den_berg_path_2009, levihn_foresight_2013, mueggler_aerial-guided_2014, castaman_sampling-based_2016, castaman_sampling-based_2016-1} do not. The most common way is to use a heuristic path planner that ignores movable obstacles to find 'blocking' obstacles \cite{stilman_navigation_2005, stilman_planning_2007, stilman_planning_2008, moghaddam_planning_2016, sun_semantic_2017, meng_active_2018}. Then, the last blocking object is selected by last intersection \cite{stilman_navigation_2005, stilman_planning_2007, stilman_planning_2008, moghaddam_planning_2016} or by least euclidean cost to go from obstacle to goal \cite{meng_active_2018}. The propositions of Wu \& Levihn \cite{wu_navigation_2010, levihn_navigation_2011, levihn_locally_2014} use a priority queue ordered by heuristic euclidean distance from obstacle to goal. Finally, the last approach is to use a graph that links obstacles to free space components so that obstacles are considered in the order they can be reached, as in Levihn \& Scholz's NAMO-MDP \cite{levihn_hierarchical_2013, levihn_planning_2013, scholz_navigation_2016} or Okada's Task Graph \cite{okada_environment_2004}.

\begin{landscape}
\begin{table}
\caption{Synthesis table with main differentiating criteria}
\label{synthesis_table}
\scalebox{0.85}{
\begin{tabular}{m{2.7cm}|x{1cm}x{1.6cm}x{2.2cm}x{1cm}x{1cm}x{1.6cm}x{1.8cm}x{2.4cm}x{1.5cm}x{1.6cm}x{1.9cm}x{1cm}}
\hline
\multicolumn{1}{c}{Reference} & Prior & Movability & Uncertainty & Comp. & Opt. & Cost & C-Space & Task P. & Transit P. & Transfer P. & Manipulations & Real-World \\ \hline
Chen \cite{chen_practical_1991} & Full & Given & None & - & - & D & Disc. & Dij. + GD & N/A & N/A & Prim. & No \\ 
Okada \cite{okada_environment_2004} & Full & Given & None & - & - & D$\|$E & Disc. & Custom & NG & NG & Grasp & No \\ 
Stilman \cite{stilman_navigation_2005} & Full & Given & None & RC & - & E+NMO & Disc. & DFS & A* & BFS & Prim. & No \\ 
Stilman \cite{stilman_planning_2007} & Full & Given & Pos. & RC & - & E+NMO & Disc. & DFS & A* & BFS & Prim. & \textbf{Yes} \\ 
Nieuwenhuisen \cite{nieuwenhuisen_effective_2008} & Full & Given & None & PC & - & D+PS & Cont. & Custom & RRT & RRT & Grasp & No \\ 
Stilman \cite{stilman_planning_2008} & Full & Given & None & - & - & D+NMO & Disc. & DFS & A* & BFS & Grasp & No \\ 
Van den Berg \cite{van_den_berg_path_2009} & Full & Given & None & PC & - & (D) & Disc. & Custom & N/A & N/A & Grasp & No \\ 
Kakiuchi \cite{kakiuchi_working_2010} & \textbf{None} & Manip. & Pos. Mov. & - & - & (D+NMO) & Cont. & Custom & RRT & N/A & Push & \textbf{Yes} \\ 
Wu \cite{wu_navigation_2010} & \textbf{None} & Manip. & None & - & - & (D$\|$T$\|$E) & Disc. & Custom & A* & DFS & Push & No \\ 
Levihn \cite{levihn_navigation_2011, levihn_locally_2014} & \textbf{None} & Manip. & None & - & \textbf{LO} & (D$\|$T$\|$E) & Disc. & Custom & D*Lite & DFS & Grasp & No \\ 
Levihn \cite{levihn_hierarchical_2013} & Full & Given & Pos. Mov. & - & - & PS & Disc. & MDP + MCTS & N/A & N/A & Prim. & No \\ 
Levihn \cite{levihn_planning_2013} & Full & Given & Pos. Mov. Kin. & - & - & T+E & Cont. & MDP + MCTS & PRM & RRT & Prim. & No \\ 
Levihn \cite{levihn_foresight_2013} & Partial & Recog. & Pos. Mov. & - & - & (D$\|$T$\|$E) & Cont. & BHPN & RRT & RRT & Grasp & \textbf{Yes} \\ 
Mueggler \cite{mueggler_aerial-guided_2014} & Full & Recog. & None & - & - & T & Disc. & Custom & A* & Dij. & Grasp & \textbf{Yes} \\ 
Castaman \cite{castaman_sampling-based_2016-1, levihn_locally_2014} & Full & Given & Pos. & - & - & T & Disc. & KPIECE + A* & N/A & N/A & Grasp$\|$Push & No \\ 
Moghaddam \cite{moghaddam_planning_2016} & Full & Given & None & \textbf{CO} & - & E & Cont. & DFS & Dij. + VG & Dij. + VG & Grasp & No \\ 
Scholz \cite{scholz_navigation_2016} & Full & Recog. & Pos. Mov. Kin. & - & - & T+E & Cont. & MDP + MCTS & PRM & RRT & Prim. & \textbf{Yes} \\ 
Sun \cite{sun_semantic_2017} & Partial & Recog. & Pos. & - & - & (D) & Cont. & Custom & RRT & RRT & Grasp$\|$Push & \textbf{Yes} \\ 
Meng \cite{meng_active_2018} & Partial & Recog. & Pos. & - & - & D & \textbf{Cont. MP} & Custom & RRT & RRT & Grasp$\|$Push & \textbf{Yes} \\ \hline
\multicolumn{13}{p{22.5cm}}{\textbf{Legend:} () = Not given but likely; '+' = Combination of; '$\|$' = Alternative to; Manip. = Found through manipulation; Recog. = Found through visual recognition; Pos. = Manage uncertainty on position; Mov. = Same on movability; Kin. = Same on object kinematics; '-' = Depending on columns, either Not Optimal or Not Complete; CO = Complete; RC = Resolution-Complete; PC = Probabilistically Complete; LO = Locally Optimal; D = Distance; E = Energy; T = Time; NMO = Number of Moved Obstacles; PS = Probability of Success; Disc. = Discrete; Cont. = Continuous; MP = Multi-Plane; Dij. = Dijkstra; GD = Generalized Distance; VG = Visibility Graph; NG = Not Given; N/A = Non Applicable; Prim. = Motion Primitives}
\end{tabular}
}
\end{table}
\end{landscape}

As for transit path planners, used ones are traditional A* \cite{stilman_navigation_2005, stilman_planning_2007, stilman_planning_2008, wu_navigation_2010, mueggler_aerial-guided_2014}, D*Lite \cite{levihn_navigation_2011, levihn_locally_2014} over discrete environments, and RRT \cite{nieuwenhuisen_effective_2008, kakiuchi_working_2010, levihn_foresight_2013, sun_semantic_2017, meng_active_2018} or PRM \cite{levihn_planning_2013, scholz_navigation_2016} variants and Dijkstra over Visibility Graph \cite{moghaddam_planning_2016} for continuous ones. On the other hand, obstacle placements are either decided through incremental application of motion primitives (forward search, eg. little translations/rotations) \cite{chen_practical_1991, stilman_navigation_2005, stilman_planning_2007, nieuwenhuisen_effective_2008,stilman_planning_2008,van_den_berg_path_2009, kakiuchi_working_2010, wu_navigation_2010, levihn_navigation_2011, levihn_hierarchical_2013, levihn_planning_2013, levihn_foresight_2013, levihn_locally_2014, scholz_navigation_2016}, or by growing sampling of possible placements in the obstacle's vicinity and subsequent path verification \cite{okada_environment_2004, mueggler_aerial-guided_2014, castaman_sampling-based_2016, castaman_sampling-based_2016-1, sun_semantic_2017, meng_active_2018}. In some cases, when planning for successive obstacles, placement is constrained by the need to keep a taboo zone for the next manipulations \cite{nieuwenhuisen_effective_2008, stilman_planning_2008, van_den_berg_path_2009, levihn_foresight_2013, moghaddam_planning_2016, sun_semantic_2017, meng_active_2018}. In the end, in discrete environments, transfer path planners iterate over possible obstacle placements using Best-First Search \cite{stilman_navigation_2005, stilman_planning_2007, stilman_planning_2008} or Depth-First Search \cite{wu_navigation_2010, levihn_navigation_2011, levihn_locally_2014}, or in continuous ones, using an RRT variant \cite{nieuwenhuisen_effective_2008, levihn_planning_2013, levihn_foresight_2013, scholz_navigation_2016, sun_semantic_2017, meng_active_2018} or again Dijkstra+VG \cite{moghaddam_planning_2016}. 

Approaches mentioned for the three planning tasks are given in Table \ref{synthesis_table}. We can note they are commonly found in SAN, thus incorporating social cost in NAMO planners should be possible. Although, many of them are offline planners: efficient online or anytime-oriented variants will be needed. In conclusion to this state of the art, we underline that none of the existing NAMO literature directly addresses social constraints, though a few references quickly mention the idea of taking object fragility into account \cite{stilman_navigation_2005, kakiuchi_working_2010}.


\section{Extension of NAMO Algorithms}\label{extension_section} 

\subsection{Objectives of Socially-Aware NAMO} 

From our previous analysis, three general objectives of S-NAMO can be identified. The first is \textbf{Social Movability Evaluation}, or determining the movability of an object by human-acceptance for a robot to move it. The second is \textbf{Social Placement Choice}, or ensuring that the final environment reconfiguration is the least disturbing to humans compared to the initial one. Finally, the third is \textbf{Social Action Planning}, or making sure that all robot actions are in themselves as safe and comfortable for humans as possible.

In the light of the classification in SAN literature \cite{kruse_human-aware_2013, rios-martinez_proxemics_2015}, we can elaborate three levels of problems of growing difficulty: \textit{delayed human-object interaction} due to future human presence, \textit{indirect interaction} due to actual human presence, and \textit{direct human-robot interaction}. At the first level, like in usual NAMO, the robot can assume to be the only autonomous agent around (eg. cleaning robot servicing while humans are away), thus it only needs to be concerned about \textbf{Social Movability Evaluation} and \textbf{Social Placement Choice}. At the next levels, the robot must also integrate the dynamic and social aspects of human presence, and answer the additional objective of \textbf{Social Action Planning}, exhibiting behaviors such as kindly asking humans to let it pass. In the rest of the paper, we make a first S-NAMO proposition addressing \textbf{Social Movability Evaluation} and \textbf{Social Placement Choice}, in the context of \textit{delayed human-object interaction}. 

\subsection{Extension of Wu and Levihn's approach} 

\SetKwFunction{algo}{make-and-execute-plan}
\SetKwFunction{subproc}{make-plan-for-obstacle}
\SetKwFunction{plan}{plan}
\SetKwFunction{path}{A*}
\SetKwFunction{update}{update-robot-knowledge}
\SetKwFunction{checkinvalid}{is-plan-invalid}
\SetKwFunction{iterate}{iterate-over-heurObsLists}
\SetKwFunction{updateml}{update-mL}
\SetKwFunction{execute}{try-execute-next-step}
\SetKwFunction{affordable}{affordable-actions}
\SetKwFunction{qfor}{q-for}
\SetKwFunction{lastlook}{get-last-look-q}
\SetKwFunction{splitatpose}{split-at-pose}
\SetKwFunction{computecs}{compute-$c_0$-$c_1$}
\SetKwFunction{ismovable}{is-movable}
\SetKwFunction{isunknown}{is-unknown}
\SetKwFunction{customcopy}{copy}
\SetKwFunction{simonestep}{sim-one-step}
\SetKwFunction{cest}{$C_{est}$}
\SetKwFunction{checknewopening}{check-new-opening}
\SetKwFunction{notintaboo}{not-in-taboo}
\SetKwFunction{cost}{cost}
\SetKwFunction{isstepsuccess}{is-step-success}
\SetKwFunction{customline}{line}
\SetKwFunction{getql}{get-qL}
\SetKwFunction{multigoalastar}{multigoal-A*}
\SetKwFunction{shortestcs}{shortest-$c_0$-$c_1$}

We chose to build our proposition upon the solution proposed by Wu \& Levihn \cite{wu_navigation_2010, levihn_navigation_2011, levihn_locally_2014} mainly for two reasons. First, it is designed for unknown environments, thus covers plan invalidation in the light of new knowledge, which is eventually essential for real-world applications. Second, as long as the problem is solvable by a single obstacle move in a single direction in the current robot knowledge, local optimality is guaranteed. It basically follows the general form of a NAMO algorithm presented in Section \ref{analysis_section}: iterate over known obstacles following a heuristic order, and evaluate potential plans that include obstacle movement as long as it can create a better plan. We introduce the S-NAMO Algorithm (see below), which extends the Wu \& Levihn approach. The algorithm relies on two procedures: a main obstacle-level one, \algo{\null}, that when needed calls a combined transfer/transit path planning sub-procedure \subproc{\null}. We first present these two procedures ignoring our S-NAMO extensions highlighted in red and blue in Algorithm \ref{snamo_algo}, then detail the extensions.

\begin{wrapfigure}[14]{L}{0.33\textwidth}
    \raisebox{0pt}[\dimexpr\height-2.0\baselineskip\relax]{
    \begin{minipage}{0.33\textwidth}
         \includegraphics[width=1.0\textwidth]{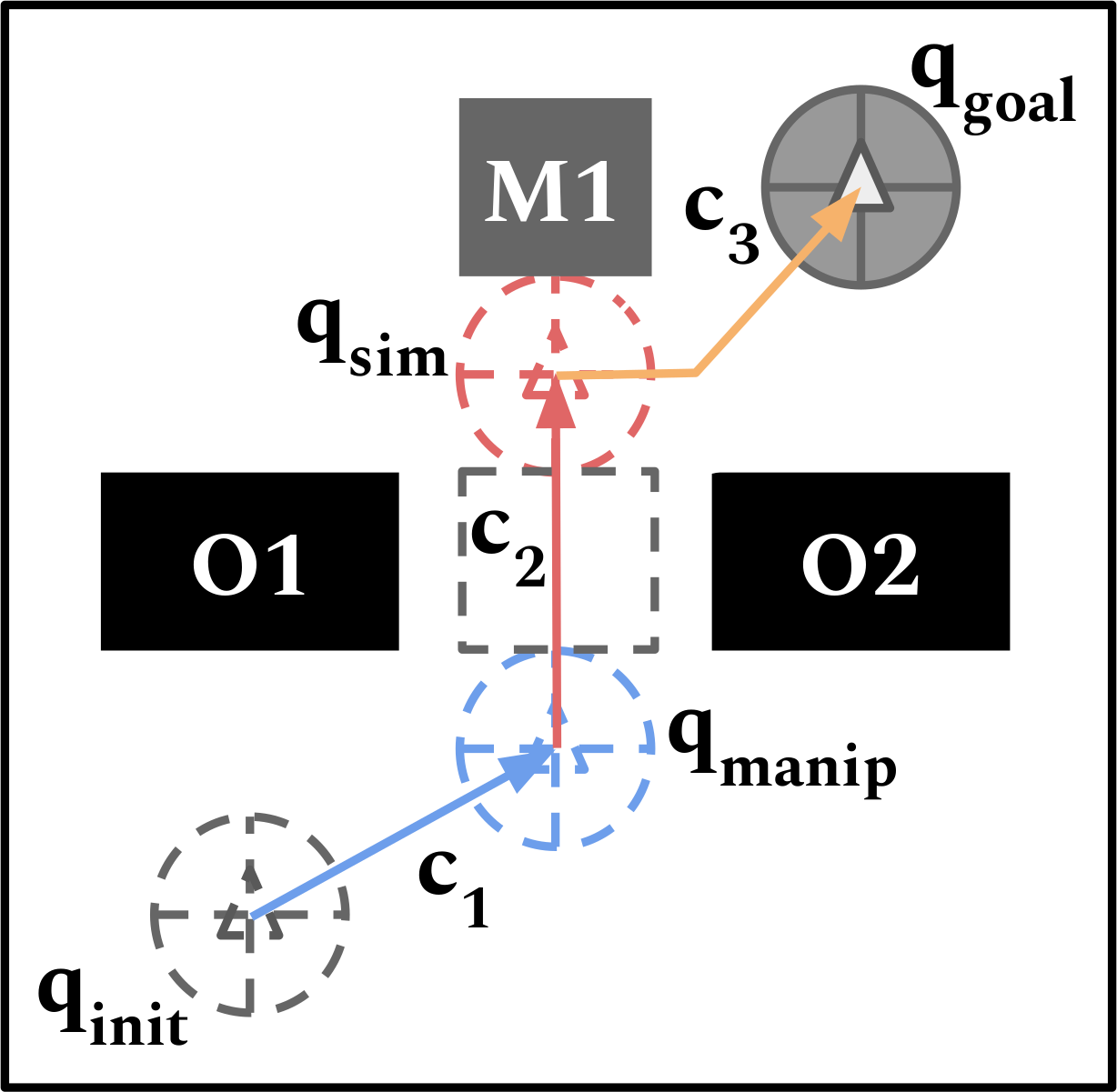}
         \caption{The robot (grey disc) executes a three-step $p_{opt}$ plan to move M1.}
         \label{path_components_fig}
    \end{minipage}
    }
\end{wrapfigure}

\vspace{3mm}

\noindent The main procedure, \algo{$w$, $q_{init}$, $q_{goal}$}, builds and executes the optimal navigation plan $p_{opt}$ from world knowledge $w$ (2D metric map with polygonal entities) and robot configurations \{$q_{init}$, $q_{goal}$\}. $p_{opt}$ is either a path avoiding all obstacles or constructed from three path components (see Figure \ref{path_components_fig}): $c_1$, $c_2$, $c_3$, respectively paths from $q_r$ to $q_{manip}$, from $q_{manip}$ to $q_{sim}$ where the robot stops moving obstacle $o$, and from $q_{sim}$ to $q_{goal}$. It always first tries to find the best plan avoiding all obstacles, and only then, iterates over movable obstacles to find out whether moving one of them will yield a better plan.
Robot knowledge is updated after each execution step, and if $p_{opt}$ is no longer valid (future collision with other obstacles by robot or manipulated obstacle, failure in manipulation, or disrupting update of the manipulated obstacle geometry), re-planning is triggered.
Since our contributions do not concern this procedure, we refer the reader to \cite{levihn_locally_2014} to better understand the iteration through obstacles.

\vspace{3mm}


\noindent The sub-procedure, \subproc{$w$, $q_{r}$, $q_{goal}$, $o$, $p_{opt}$}, is called during the iteration over obstacles, and returns the best plan $p_{best}$ implying the manipulation of obstacle $o$. It iterates over actions $act$ that can be done on $o$, assuming (line \ref{qforline}) there is only one robot configuration $q_{manip}$ for every \{$o$, $act$\} pair (middle of $o$'s side). The plan components are computed sequentially, starting with $c_1$. If $c_1$ is found, successive unit actions $act$ of constant length are simulated ($count$ times) in a copy of $w$ until impossible (collision with other obstacle). To avoid unnecessary computations of $c_2$ and $c_3$, the simulation is stopped as soon as an underestimated cost $C_{est}$ of the currently evaluated plan gets higher than the one of $p_{opt}$ (l.\ref{cestline}). $C_{est}$ is the sum of $c_1$'s cost, a $c_2$ estimate (product of $count$ by unit length), and a $c_3$ estimate (minimal euclidean distance between $o$ and $q_{goal}$). Also, full evaluation is only done if a new local opening has been created around $o$ (l.\ref{taboo_line}, method described in \cite{levihn_efficient_2011}).

\begin{figure}
\vspace*{-0.2in}
\centering
\captionsetup[subfigure]{justification=centering}
\begin{subfigure}{0.28\textwidth}
  \centering
  \includegraphics[width=\linewidth]{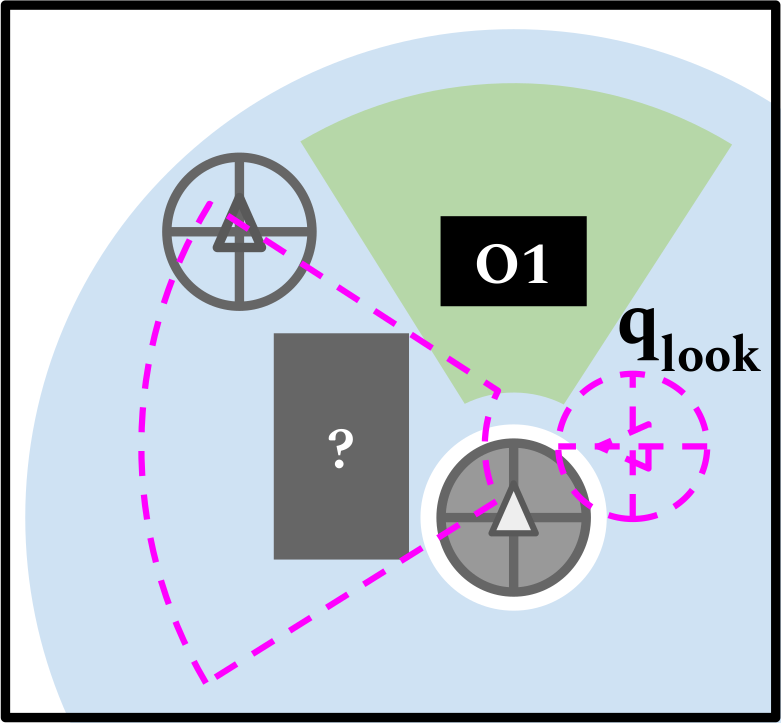}
  \caption{}
  \label{fig:obs_1}
\end{subfigure}
\begin{subfigure}{0.28\textwidth}
  \centering
  \includegraphics[width=\linewidth]{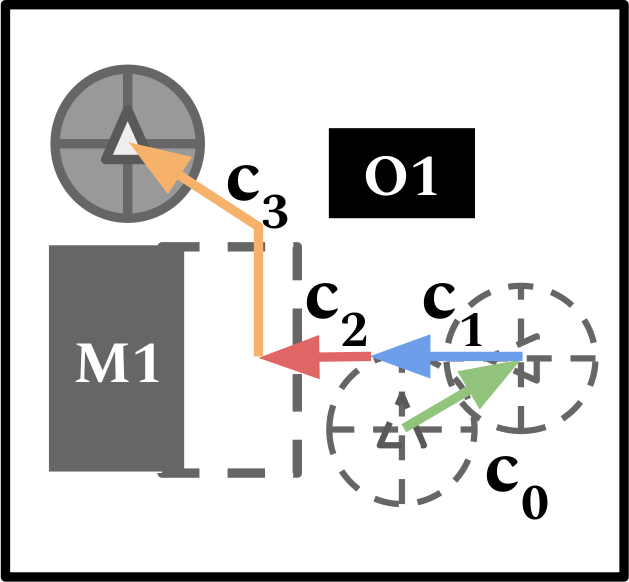}
  \caption{}
  \label{fig:obs_2}
\end{subfigure}
\begin{subfigure}{0.20\textwidth}
  \centering
  \includegraphics[width=\linewidth]{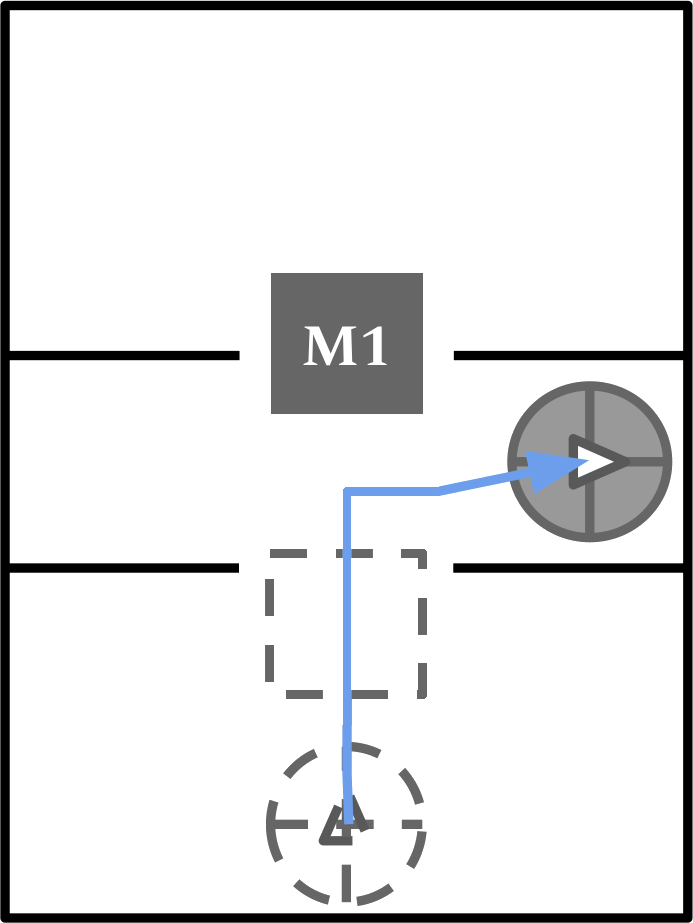}
  \caption{}
  \label{fig:taboo_1}
\end{subfigure}
\begin{subfigure}{0.20\textwidth}
  \centering
  \includegraphics[width=\linewidth]{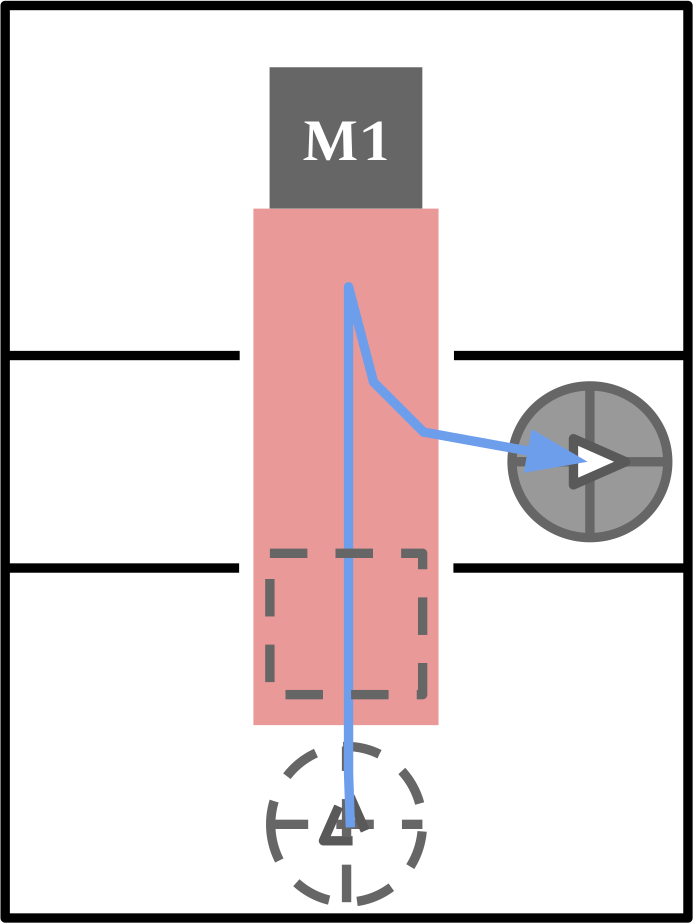}
  \caption{}
  \label{fig:taboo_2}
\end{subfigure}
\caption{\textbf{In (a)}, G-FOV (blue) detected two obstacles, S-FOV (green) only identified unmovable obstacle O1. Robot is too close from other obstacle to observe it. Going through best intermediate observation configuration is necessary: final best path with $c_0$ is shown in (b). \textbf{(c)} represents two facing rooms separated by a corridor. In typical NAMO (c), robot will push M1 just enough to pass, blocking the other doorway. In our S-NAMO proposition, the taboo zone (red) prevents blocking, but may end up with a longer plan.}
\label{fig:propal_illus}
\vspace*{-0.3in}
\end{figure}

\begin{algorithm}[H]
	\label{snamo_algo}
    \caption{\textbf{S-NAMO} - Extension of the Wu\&Levihn approach: Social Movability Evaluation in \color{blue}blue \color{black} and Social Placement Choice in \color{red}red\color{black}}

    \DontPrintSemicolon
    \scriptsize
    \footnotesize    
    \SetKwProg{makeexe}{Procedure}{}{}
    \SetKwProg{planforobs}{Procedure}{}{}
    \SetKwProg{compute}{Procedure}{}{}
    
    \makeexe{\algo{$w$, $q_{init}$, $q_{goal}$}}{
        $\rhd$ when plan is invalidated, makes a plan avoiding all obstacles, then tries to improve it by iterating over obstacles and calling \subproc{$w$, $q_r$, $q_{goal}$, $o$, $p_{opt}$}\;
    }{}
    
    
    \setcounter{AlgoLine}{0}
    
    \planforobs{\subproc{$w$, $q_{r}$, $q_{goal}$, $o$, $p_{opt}$}}{
        $p_{best} \gets \emptyset$\;
        \ForEach{$act$ {\normalfont\textbf{in}} \affordable{$o$}}{
            $q_{manip}, c_1 \gets$ \qfor{$o$, $act$}, \path{$w$, $q_r$, $q_{manip}$}\; 
            \label{qforline}
            \If{$c_1 \neq \emptyset$}{
                \color{blue}
                \If{\isunknown{$o$}}{ \label{unknown_check_line}
                    $q_{look} \gets$ \lastlook{$w$, $o$, $c_1$}\; \label{c1_obs_line}
                    \lIf{$q_{look} \neq \emptyset$}{$c_0, c_1 \gets$ \splitatpose{$c_1$, $q_{look}$}}
                    \lElse{$c_0, c_1 \gets$ \computecs{$w$, $o$, $q_r$, $q_{manip}$}} \label{compute_c0_c1_line}
                }
                \lElse{$c_0 \gets \emptyset$}
                \color{black}
            
                \If{\color{blue}\ismovable{$o$} \textbf{or} (\isunknown{$o$} \textbf{and} $c_0 \neq \emptyset$ \textbf{and} $c_1 \neq \emptyset$)\color{black}}{
                    $w_{sim} \gets$ \customcopy{$w$}\;
                    $count, q_{sim} \gets$ 1, \simonestep{$w_{sim}$, $act$, $o$, $q_r$}\;
                    \While{\cest{$w_{sim}$, $c_1$, $count$, $act$} $\leq$ \cost{$p_{opt}$} \\ \ \textbf{and} \isstepsuccess{$q_r$, $q_{sim}$, $count$, $act$}}{ \label{cestline}
                        \If{\checknewopening{$w$, $w_{sim}$, $o$} \color{red}\textbf{and} \notintaboo{$w$, $o$}\color{black}}{ \label{taboo_line}
                            $c_2 \gets$ \customline{$q_{manip}$, $q_{sim}$}\;
                            $c_3 \gets$ \path{$w$, $q_{sim}$, $q_{goal}$}\;
                            \If{$c_3 \neq \emptyset$}{
                                $p \gets$ \plan{\color{blue}$c_0$, \color{black}$c_1$, $c_2$, $c_3$, $o$, $act$}\;
                                \lIf{\cost{$p$} $<$ \cost{$p_{best}$}}{$p_{best} \gets p$}
                                \lIf{\cost{$p_{best}$} $<$ \cost{$p_{opt}$}}{$p_{opt} \gets p_{best}$}
                            }
                        }
                        $count, q_{sim} \gets$ $count + 1$, \simonestep{$w_{sim}$, $act$, $o$, $q_{sim}$}\;
                    }
                }
            }
        }
        \KwRet $p_{best}$ \;
    }
    
    \setcounter{AlgoLine}{0}
    
    \color{blue}
    \compute{\computecs{$w$, $o$, $q_r$, $q_{manip}$}}{
        $qL \gets$ \getql{$w$, $o$}\; \label{get_ql_line}
        $paths$-$qL$-$q_{manip} \gets$ \multigoalastar{$w$, $q_{manip}$, $qL$}\; \label{first_multigoal_astar_line}
        $paths$-$q_r$-$qL \gets$ \multigoalastar{$w$, $q_r$, $qL$}\; \label{second_multigoal_astar_line}
        \KwRet \shortestcs{$paths$-$q_r$-$qL$, $paths$-$qL$-$q_{manip}$}\; \label{return_best_line}
	}
	\color{black}
\end{algorithm}
\smallskip
\noindent \textbf{Social Movability Evaluation} The initial approach of Wu \& Levihn supposes that any obstacle is movable unless a manipulation tentative failed, in which case it is blacklisted. However, in S-NAMO this is not an acceptable behavior since it could lead to unauthorized objects manipulations. As a first approach, we propose a simple white-listing system: unregistered obstacles are considered unmovable. But a robot often relies on multiple sensors, and their respective Fields Of View (FOV) are not necessarily equal. An obstacle may have been detected geometrically, but not yet identified, leading to three possible states: unknown, movable, unmovable. As in the initial algorithm, we suppose a perfect conical `geometry sensor' (eg. high resolution laser range finder), with perfect segmentation of obstacles (blue disk in Figure \ref{fig:obs_1}), but we add a perfect `semantic sensor' that guarantees identification if the obstacle is in its FOV (eg. using a RGB-D Camera). The geometric FOV (G-FOV) is assumed to cover more space than the semantic one (S-FOV). White-listed obstacles are assumed to fit into the S-FOV, anything that doesn't is automatically classed as unmovable.

The \subproc{\null} procedure has been adapted to work under these hypotheses. When obstacle $o$ is known as movable, the algorithm is unchanged. When $o$ is unknown, we first check whether the usual computation of $c_1$ can provide observation certainty (lines \ref{unknown_check_line}-\ref{c1_obs_line}); if not, we try to find another path that guarantees observation (line \ref{compute_c0_c1_line}). To do that, in \computecs{\null}, we determine the discrete robot configurations list $qL$ that would allow observation (l.\ref{get_ql_line}): first, we get all non-colliding configurations within the area between the inflated obstacle polygons by minimal and maximal observation distances, and among them we only return these where $o$ is included in the S-FOV. Then, we execute the multi-goal A* algorithm between $q_{manip}$ and every configuration in $qL$ (l.\ref{first_multigoal_astar_line}). The same is done from the current robot configuration $q_r$ to all elements of $qL$ (l.\ref{second_multigoal_astar_line}). Finally, we return the best pair of paths \{$c_0$, $c_1$\} (l.\ref{return_best_line}): see illustration Fig. \ref{fig:obs_2}.



\smallskip
\noindent \textbf{Social Placement Choice} In current NAMO approaches, the robot does not care about placing obstacles in socially-critical spots (eg. around doors, often-used furniture, \dots). As a first step, we answer this problem with a binary approach: either the zone is taboo for obstacle placement, or it is not.

We extend the definition of $w$ by adding a social placement semantic map layer, where
taboo zones are defined as a set of polygons $P$.
We assume for now that $P$ is provided by human users. Now, whenever the polygonal footprint of an obstacle intersects with any polygonal taboo zone in $P$, \notintaboo{$w$, $o$} returns $False$, preventing full plan evaluation (see \subproc{\null} procedure, line \ref{taboo_line}). Figure \ref{fig:taboo_2}) illustrates this process on a simple scenario. Next section presents experiments with more complex scenarios.

\section{Experiments}

\label{experimentations_section}  

We implemented the S-NAMO algorithm in a custom simulator based on ROS standards. This is a first step toward an implementation on a real robot (Pepper), simplifying object detection and identification, which could later be addressed with an existing package such as \href{https://github.com/tue-robotics/ed}{ED} from TU-Eindhoven. For the sake of implementation ease, movable obstacles are assumed to be convex polygons. All computations are done on the 2D vectorial model, except for path planning, which is implemented as a grid-search A* Algorithm, as in \cite{wu_navigation_2010} \footnote{\label{code_footnote}All the code and its execution instructions are available on the following repository: \url{https://gitlab.inria.fr/brenault/s-namo-sim}}.


\vspace{2mm}

\begin{figure}
\vspace*{-0.2in}
\centering
\captionsetup[subfigure]{justification=centering}
\begin{subfigure}{0.193\textwidth}
  \centering
  \includegraphics[width=\linewidth]{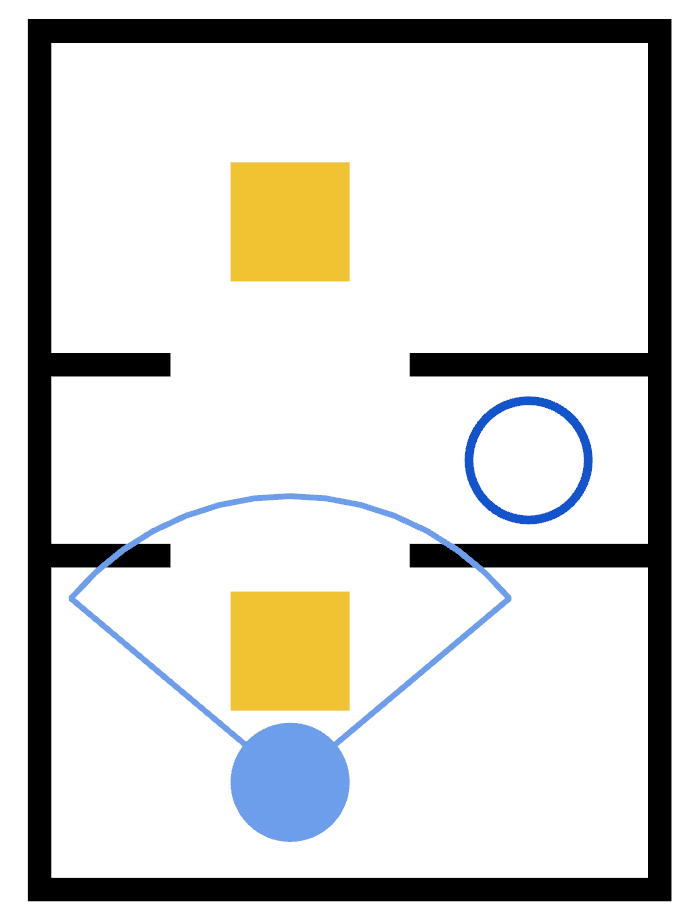}
  \caption{}
  \label{fig:sub1}
\end{subfigure}
\begin{subfigure}{0.193\textwidth}
  \centering
  \includegraphics[width=\linewidth]{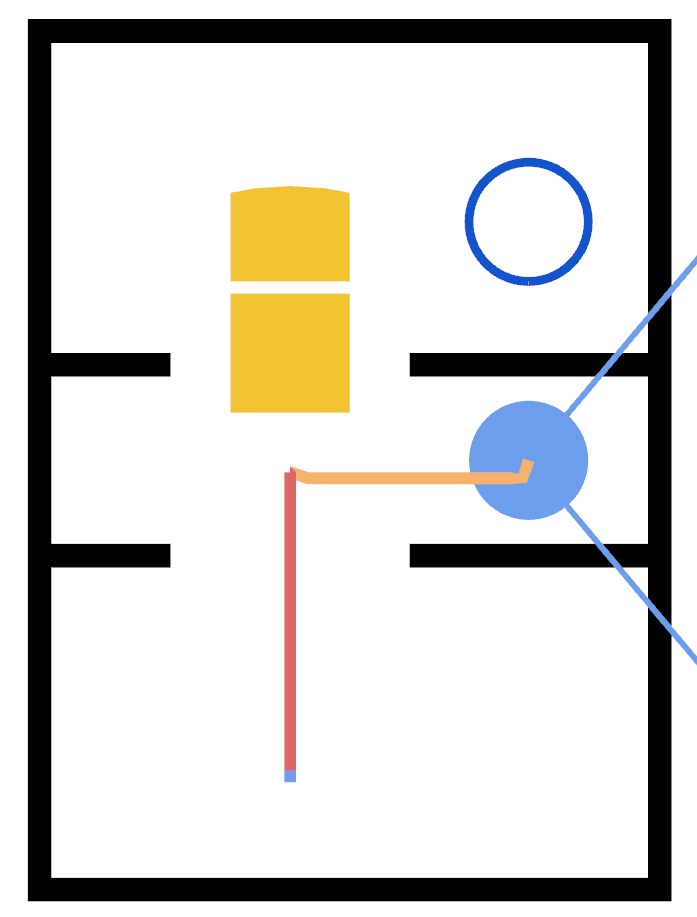}
  \caption{}
  \label{fig:sub2}
\end{subfigure}
\begin{subfigure}{0.193\textwidth}
  \centering
  \includegraphics[width=\linewidth]{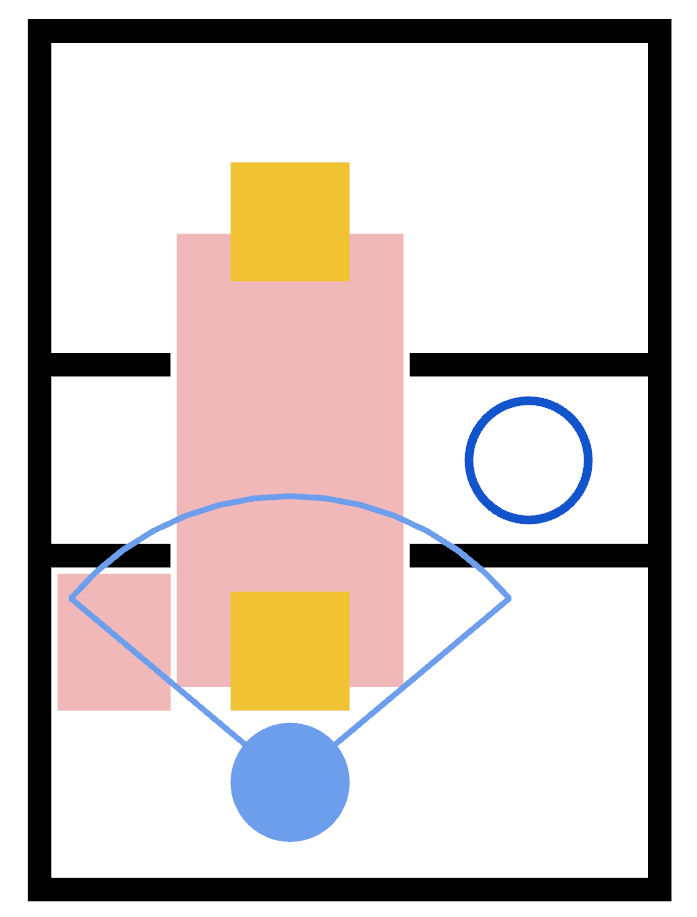}
  \caption{}
  \label{fig:sub3}
\end{subfigure}
\begin{subfigure}{0.193\textwidth}
  \centering
  \includegraphics[width=\linewidth]{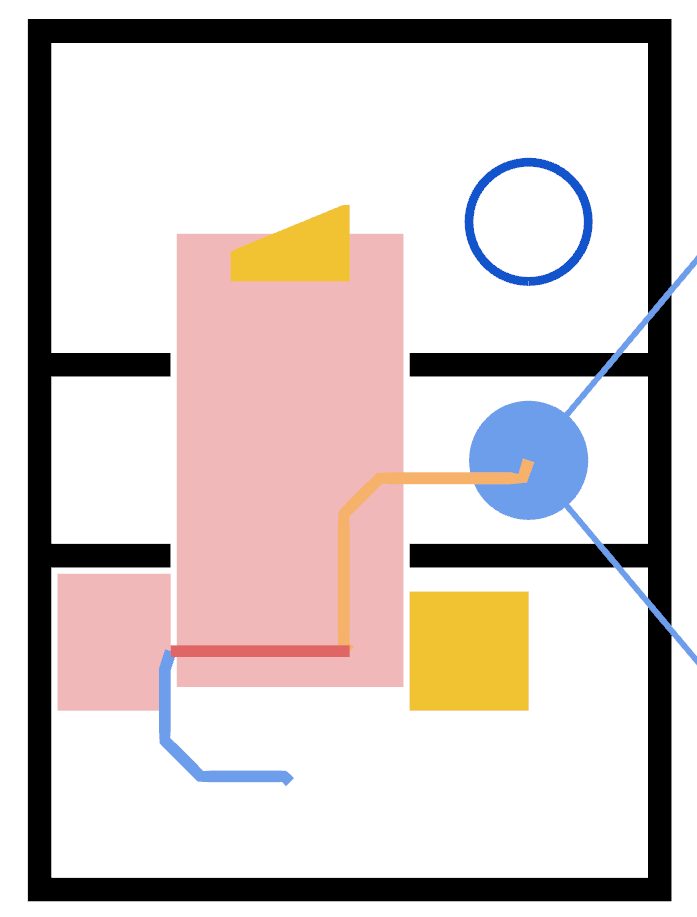}
  \caption{}
  \label{fig:sub4}
\end{subfigure}
\begin{subfigure}{0.193\textwidth}
  \centering
  \includegraphics[width=\linewidth]{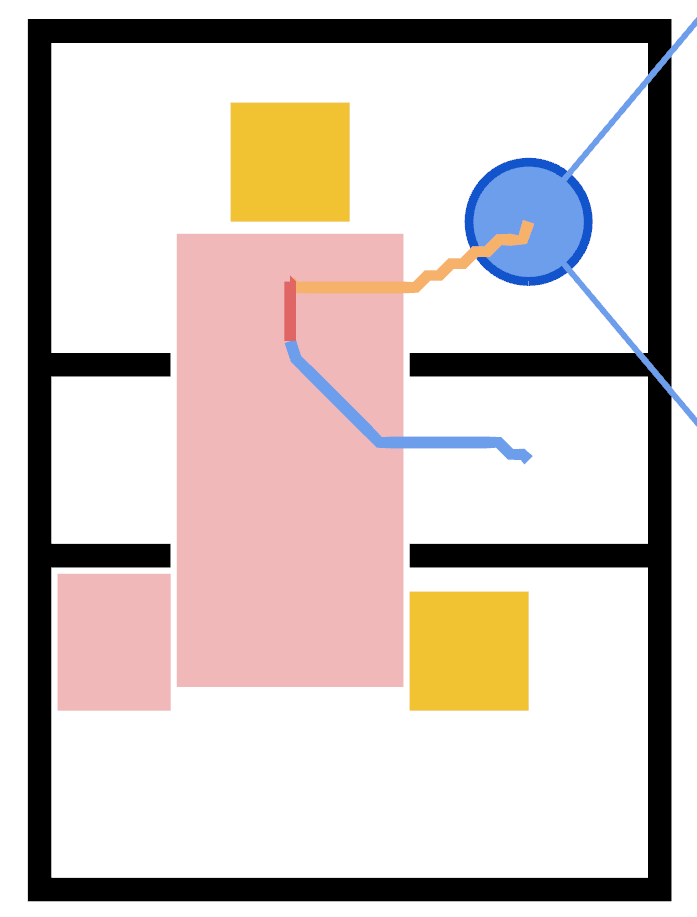}
  \caption{}
  \label{fig:sub5}
\end{subfigure}
\vspace*{-0.08in}
\caption{Simulation of a two-goals scenario with NAMO (a,b) vs. S-NAMO (c,d,e) 
}
\label{fig:test}
\vspace*{-0.3in}
\end{figure}

\noindent \textbf{Social Placement Choice} We tested the Social Placement Choice process in a scenario where a robot has to successively reach two goals represented as empty circles in Fig.\ref{fig:test}. The environment consists of two rooms separated by a corridor, but two yellow boxes are blocking the doorways (Fig.\ref{fig:sub1}). The robot (blue circle, FOV is the cone) starts from the bottom room.

In Fig.\ref{fig:sub2} we see that a standard NAMO approach like Wu \& Levihn's results in blocking the other doorway: only the first goal is reached. In Fig.\ref{fig:sub3} we introduce the social semantic layer which consists in two taboo areas for objects (in red), respectively related to the doorways and a `precious carpet'. Fig.\ref{fig:sub4} shows how our algorithm deals with the first discovered object: the computed path moves the box on the right and outside taboo areas, leaving in particular the doorway area free. In Fig.\ref{fig:sub5}, the robot encountered the second box and pushed it outside the taboo area, leading to a path reaching the second goal. This S-NAMO scenario (and others) are available as videos \footnote{\label{video_footnote}Link to videos : \url{https://gitlab.inria.fr/brenault/s-namo-sim/wikis/Videos}}.

\begin{figure}
    \begin{minipage}{\linewidth}
        \centering\captionsetup[subfigure]{justification=centering}
        \begin{subfigure}{0.5\textwidth}
          \centering
          \includegraphics[width=\linewidth]{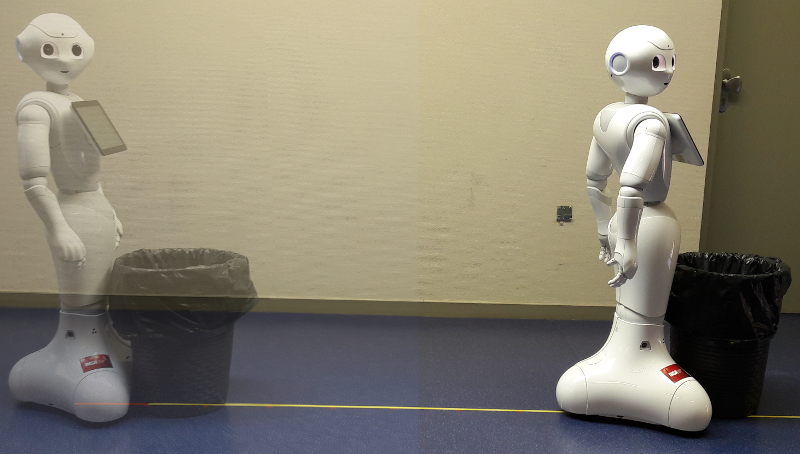}
          \caption{}
          \label{fig:pepper_01}
        \end{subfigure}
        \begin{subfigure}{0.15\textwidth}
          \centering
          \includegraphics[width=\linewidth]{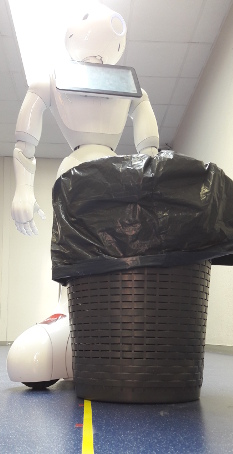}
          \caption{}
          \label{fig:pepper_02}
        \end{subfigure}
        \begin{subfigure}{0.15\textwidth}
          \centering
          \includegraphics[width=\linewidth]{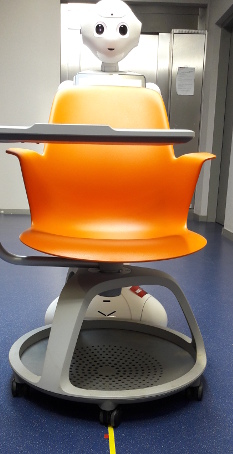}
          \caption{}
          \label{fig:pepper_03}
        \end{subfigure}
            \begin{subfigure}{0.15\textwidth}
          \centering
          \includegraphics[width=\linewidth]{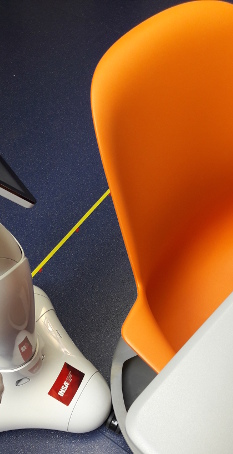}
          \caption{}
          \label{fig:pepper_04}
        \end{subfigure}
    \end{minipage}
    \vspace*{-0.08in}
\captionsetup{justification=centering}
\caption{Pepper pushing a bin and a chair}\label{fig:pepper_push}
\end{figure}

\noindent \textbf{Pushing Experiments with Pepper} In the view of a real-world implementation, we experimented with Pepper's base pushing abilities, using our existing robot architecture developed for the Robocup@Home 2018 \cite{jumel_context_2018}. In Fig.\ref{fig:pepper_01} and \ref{fig:pepper_02}, Pepper successfully pushes a garbage bin in a straight line with little deviation. We have also verified that with other light objects such as cardboard boxes, that when the object's side is properly centered relatively to the robot, pushes are more likely to succeed. However, we also learned that, as seen in Fig.\ref{fig:pepper_03} and \ref{fig:pepper_04}, heavier objects of interest such as chairs with wheel-casters will need to be accompanied with the arms in some way to avoid unpredictable drift, but even so, the manipulation could still fail (videos available at footnote \ref{video_footnote}). Thus, in our future work, we will also strive to address uncertainty as to manipulation success, like in \cite{levihn_planning_2013}.



\section{Conclusion}\label{conclusion_section}  


In this paper, we first analyzed existing NAMO approaches in order to adapt them to social constraints. This led us to extend the Wu\&Levihn approach, by defining the S-NAMO algorithm which introduces Social Movability Evaluation and Social Placement Choice for object manipulations. We implemented these propositions in an open source ROS compatible simulator. 
Experiments showed how social semantic areas can prevent obstruction of places like circulation zones, and how the robot can identify obstacles to compute its plan.
In future works, we plan to refine the semantic layer and address actual human presence with indirect or direct human-robot interaction, while integrating ways to manage uncertainty as to sensor data or success of manipulation. We will continue to experiment and validate these social NAMO abilities with robots such as Pepper and demonstrate their interest in the RoboCup@Home challenge.



\newpage

\bibliographystyle{unsrt_mod}
\bibliography{refs}

\end{document}